\documentclass{article}
\usepackage{spconf,amsmath,graphicx}

\usepackage{algorithm}
\usepackage{algorithmic}
\usepackage{multirow}
\usepackage{multicol}
\usepackage{float}
\usepackage{booktabs}

\newcommand{\alghere}{
\begin{algorithm}[H]
\caption{Token-wise beam search for RNN-T. Subscript $h$ is added to variables to indicate the hypothesis index when applicable. See text in Section \ref{sec:method} for a high level description of the algorithm and details about the helper functions used.} 
\label{alg:alg1} 
\begin{algorithmic} 
\STATE Inputs: 
\STATE \, \, $enc$: a sequence of encoder output frames
\STATE \, \, $N$: beam size
\STATE \, \, $S$: decoding segment size
\STATE \# Hypotheses have members: tok: tokens, s: score,
\STATE \# o: predictor output, d: last token frame distribution.
\STATE B = [([], $1$, $\mathtt{InitPredictor}()$, Null)]
\STATE t = 1
\WHILE{$t \leq \mathtt{Length}(enc)$}

    \STATE $seg = enc[t : t + S]$
    \STATE $A = \mathtt{ChooseNBest}(B)$
    \STATE $B = []$
    \STATE $t = t + S$
    \FOR{$h$ from $1$ to $\mathtt{Length}(A)$}
        \STATE $A_h.\mathtt{d} = (A_h.\mathtt{s}, 0, ..., 0)$
    \ENDFOR
    \WHILE{$\mathtt{Length}(A) > 0$}
        
        \STATE $pred = [A_h.\mathtt{o} \text{ for } h \text{ from } 1 \text{ to }  \mathtt{Length}(A)]$
        \STATE $J = \mathtt{Softmax}(\mathtt{Joiner}(seg, pred))$
        \STATE

        \STATE \# The block below follows Eq. \ref{eq:eq3}-\ref{eq:eq7}, which compute \STATE \# the expansion scores for the different tokens
        \STATE $d\_old = [A_h.d  \text{ for } h \text{ from } 1 \text{ to }  \mathtt{Length}(A)]$
        \STATE $Bl = \mathtt{BlankProduct}(J)$
        \STATE $d = \mathtt{NonBlankExpansionT}(J, Bl, d\_old)$
        \STATE $\delta^{nb} = \mathtt{NonBlankExpansion}(d)$
        \STATE $\delta^{\phi} = \mathtt{BlankExpansion}(Bl, d\_old)$
        \STATE
        
        \STATE \# Add all blank expansions to $B$ and merge if needed
        \FOR{$h$ from $1$ to $\mathtt{Length}(A)$}
            \STATE $new\_hypo = (A_h.tok, \delta^{\phi}_h, A_h.o, \mathtt{Null})$
            \STATE $\mathtt{AddAndMerge}(B, new\_hypo)$
        \ENDFOR
        \STATE
        
        \STATE \# Choose up to $N$-best non-blank expansions and 
        \STATE \# add to $A$
        \STATE $A' = A$
        \STATE $A = []$
        \STATE $threshold = \mathtt{ChooseNthScore}(B, N)$
        \FOR{$(h, k)$ in $\mathtt{ChooseNBestExpansions}(\delta^{nb}$)}
            \STATE $new\_hypo = (A'_h.tok + k, \delta^{nb}_h(k), \mathtt{Null}, d_h(k))$
            \IF{$\mathtt{Length}(B) < N$ or $\delta^{nb}_h(k) > threshold$}
                \STATE $A.\mathtt{Append}(new\_hypo)$
            \ENDIF
        \ENDFOR
        \STATE $\mathtt{UpdatePredictorOutput}(A)$
    \ENDWHILE
\ENDWHILE
\RETURN $\mathtt{ChooseNBest}(B)$
\end{algorithmic}
\end{algorithm}
}

\newcommand{\tabresone}{
\begin{table*}[htb]
  \caption{Librispeech model (test-other): decoding results with the standard algorithm (segment size 1) and our proposed token-wise beam search. In parentheses are the relative improvement \% over the standard algorithm. Best throughput and OWER values are in boldface. \vspace{0.1cm}}
  \label{tab:tab1}
  \centering
  \begin{tabular}{ c c c c c c c c}
    \toprule
        N-best & Segment & WER & OWER & Throughput  & Throughput & Calls/ & Joins/ \\
               & Size    &     &      &  CPU (f/s)  & GPU (f/s)  & Frame  & Frame \\
        
\midrule     
\multirow{7}{*}{2} & 
  1 & 8.66 & 8.15 & 192.12 & 439.56 & 1.41 & 1.41 \\
& 2 & 8.67 (-0.12\%) & 7.82 (4.05\%) & 220.27 (14.65\%) & 685.14 (55.87\%) & 0.75 & 1.50 \\
& 3 & 8.63 (0.35\%) & 7.69 (5.64\%) &\textbf{ 304.62 (58.56\%)} & \textbf{861.02 (95.88\%)} & 0.54 & 1.62 \\
& 5 & 8.62 (0.46\%) & 7.51 (7.85\%) & 200.64 (4.43\%) & 770.39 (75.26\%) & 0.38 & 1.90 \\
& 10 & 8.61 (0.58\%) & 7.41 (9.08\%) & 191.48 (-0.33\%) & 392.34 (-10.74\%) & 0.27 & 2.61 \\
& 20 & 8.59 (0.81\%) & 7.38 (9.45\%) & 114.41 (-40.45\%) & 133.56 (-69.62\%) & 0.20 & 3.91 \\
& 50 & 8.57 (1.04\%) & \textbf{7.27 (10.80\%)} & 37.80 (-80.32\%) & 35.74 (-91.87\%) & 0.16 & 7.56 \\

\midrule     
\multirow{7}{*}{5} & 
  1 & 8.64 & 6.39 & 90.54 & 298.30 & 1.58 & 1.58 \\
& 2 & 8.64 (-0.00\%) & 6.19 (3.13\%) & 109.49 (20.93\%) & \textbf{455.42 (52.67\%)} & 0.83 & 1.65 \\
& 3 & 8.6 (0.46\%) & 6.11 (4.38\%) & 127.37 (40.68\%) & 447.58 (50.04\%) & 0.60 & 1.78 \\
& 5 & 8.58 (0.69\%) & 5.95 (6.89\%) & \textbf{137.15 (51.48\%)} & 370.06 (24.06\%) & 0.42 & 2.07 \\
& 10 & 8.6 (0.46\%) & 5.86 (8.29\%) & 93.57 (3.35\%) & 142.18 (-52.34\%) & 0.29 & 2.81 \\
& 20 & 8.58 (0.69\%) & 5.8 (9.23\%) & 52.34 (-42.19\%) & 61.06 (-79.53\%) & 0.21 & 4.16 \\
& 50 & 8.55 (1.04\%) & \textbf{5.76 (9.86\%)} & 15.70 (-82.66\%) & 12.83 (-95.70\%) & 0.17 & 7.92 \\

\midrule     
\multirow{7}{*}{10} & 
  1 & 8.58 & 5.42 & 48.40 & 202.95 & 1.69 & 1.69 \\
& 2 & 8.57 (0.12\%) & 5.26 (2.95\%) & 67.44 (39.34\%) & 284.28 (40.07\%) & 0.89 & 1.77 \\
& 3 & 8.59 (-0.12\%) & 5.21 (3.87\%) & \textbf{82.84 (71.16\%)} & \textbf{299.22 (47.44\%)} & 0.64 & 1.90 \\
& 5 & 8.56 (0.23\%) & 5.16 (4.80\%) & 71.86 (48.47\%) & 187.29 (-7.72\%) & 0.45 & 2.21 \\
& 10 & 8.56 (0.23\%) & 5.1 (5.90\%) & 50.65 (4.65\%) & 77.13 (-62.00\%) & 0.30 & 2.95 \\
& 20 & 8.57 (0.12\%) & 5.07 (6.46\%) & 30.33 (-37.33\%) & 33.22 (-83.63\%) & 0.22 & 4.34 \\
& 50 & 8.57 (0.12\%) & \textbf{5 (7.75\%)} & 7.35 (-84.81\%) & 7.30 (-96.40\%) & 0.18 & 8.20 \\

\bottomrule
    \end{tabular}
\end{table*}
}

\newcommand{\tabrestwo}{
\begin{table*}[htb]
  \caption{Large assistant model: decoding results with the standard algorithm (segment size 1) and our proposed token-wise beam search. In parentheses are the relative improvement \% over the standard algorithm. Best throughput and OWER values are in boldface. \vspace{0.1cm}}
  \label{tab:tab2}
  \centering
  \begin{tabular}{ c c c c c c c c}
    \toprule
        N-best & Segment & WER & OWER & Throughput  & Throughput & Calls/ & Joins/ \\
               & Size    &     &      &  CPU (f/s)  & GPU (f/s)  & Frame  & Frame \\
        
\midrule     
\multirow{7}{*}{2} &
  1 & 8.71 & 7.48 & 244.91 & 618.23 & 1.28 & 1.28 \\
& 2 & 8.56 (1.72\%) & 7.3 (2.41\%) & 241.64 (-1.34\%) & 859.12 (38.96\%) & 0.74 & 1.46 \\
& 3 & 8.49 (2.53\%) & 7.22 (3.48\%) & 348.79 (42.42\%) & 978.77 (58.32\%) & 0.54 & 1.61 \\
& 5 & 8.41 (3.44\%) & 7.04 (5.88\%) & \textbf{350.63 (43.17\%)} & \textbf{1,158.79 (87.44\%)} & 0.35 & 1.71 \\
& 10 & 8.43 (3.21\%) & 6.98 (6.68\%) & 306.70 (25.23\%) & 702.09 (13.56\%) & 0.21 & 2.02 \\
& 20 & 8.35 (4.13\%) & 6.97 (6.82\%) & 221.90 (-9.40\%) & 275.44 (-55.45\%) & 0.13 & 2.54 \\
& 50 & 8.32 (4.48\%) &\textbf{ 6.82 (8.82\%)} & 94.19 (-61.54\%) & 102.50 (-83.42\%) & 0.09 & 3.61 \\

\midrule 
\multirow{7}{*}{5} & 
  1 & 8.39 & 5.47 & 105.16 & 330.69 & 1.57 & 1.57 \\
& 2 & 8.35 (0.48\%) & 5.38 (1.65\%) & 128.45 (22.15\%) & 459.99 (39.10\%) & 0.91 & 1.81 \\
& 3 & 8.31 (0.95\%) & 5.31 (2.93\%) & \textbf{141.48 (34.54\%)} & 515.16 (55.78\%) & 0.68 & 2.01 \\
& 5 & 8.29 (1.19\%) & 5.27 (3.66\%) & 136.85 (30.14\%) & \textbf{537.73 (62.61\%)} & 0.43 & 2.13 \\
& 10 & 8.29 (1.19\%) & 5.19 (5.12\%) & 127.37 (21.12\%) & 215.88 (-34.72\%) & 0.26 & 2.51 \\
& 20 & 8.23 (1.91\%) & 5.17 (5.48\%) & 94.33 (-10.30\%) & 110.68 (-66.53\%) & 0.17 & 3.14 \\
& 50 & 8.27 (1.43\%) & \textbf{5.15 (5.85\%)} & 37.43 (-64.41\%) & 40.59 (-87.73\%) & 0.10 & 4.28 \\

\midrule     
\multirow{7}{*}{10} &
  1 & 8.25 & 4.37 & 59.65 & 212.85 & 1.83 & 1.83 \\
& 2 & 8.26 (-0.12\%) & 4.35 (0.46\%)  & 77.69 (30.24\%) & 286.10 (34.41\%) & 1.05 & 2.09 \\
& 3 & 8.27 (-0.24\%) & 4.35 (0.46\%) & \textbf{90.58 (51.85\%)} & \textbf{307.55 (44.49\%)} & 0.79 & 2.33 \\
& 5 & 8.29 (-0.48\%) & 4.29 (1.83\%) & 88.86 (48.97\%) & 203.29 (-4.49\%) & 0.50 & 2.47 \\
& 10 & 8.26 (-0.12\%) & 4.26 (2.52\%)& 74.98 (25.70\%) & 112.85 (-46.98\%) & 0.30 & 2.89 \\
& 20 & 8.23 (0.24\%) & 4.24 (2.97\%) & 48.71 (-18.34\%) & 58.18 (-72.67\%) & 0.19 & 3.58 \\
& 50 & 8.24 (0.12\%) & \textbf{4.2 (3.89\%)} & 20.18 (-66.17\%) & 19.21 (-90.97\%) & 0.11 & 4.79 \\

\bottomrule
    \end{tabular}
\vspace{-0.3cm}
\end{table*}
}

\newcommand{\tabresthree}{
\begin{table*}[hbt]
  \caption{Small dictation model: Decoding results with the standard algorithm (segment size 1) and our proposed token-wise beam search. In parentheses are the relative improvement \% over the standard algorithm. Best throughput and OWER values are in boldface. \vspace{0.1cm}} 
  \label{tab:tab3}
  \centering
  \begin{tabular}{ c c c c c c c c}
    \toprule
        N-best & Segment & WER & OWER & Throughput  & Throughput & Calls/ & Joins/ \\
               & Size    &     &      &  CPU (f/s)  & GPU (f/s)  & Frame  & Frame \\
        
\midrule     
\multirow{7}{*}{2} &
  1 & 4.14 & 3.71 & 464.99 & 807.43 & 0.94 & 0.94 \\
& 2 & 4.16 (-0.48\%) & 3.57 (3.77\%)  & 599.36 (28.90\%) & 1,289.84 (59.75\%) & 0.53 & 1.06 \\
& 3 & 4.12 (0.48\%) & 3.5 (5.66\%) & \textbf{683.26 (46.94\%)} & 1,474.57 (82.63\%) & 0.39 & 1.16 \\
& 5 & 4.14 (-0.00\%) & 3.49 (5.93\%) & 484.32 (4.16\%) & \textbf{ 1,532.11 (89.75\%)} & 0.28 & 1.38 \\
& 10 & 4.13 (0.24\%) & 3.45 (7.01\%) & 457.11 (-1.69\%) & 573.59 (-28.96\%) & 0.19 & 1.85 \\
& 20 & 4.12 (0.48\%) & 3.42 (7.82\%) & 156.37 (-66.37\%) & 222.66 (-72.42\%) & 0.14 & 2.74 \\
& 50 & 4.14 (-0.00\%) & \textbf{3.39 (8.63\%)} & 72.45 (-84.42\%) & 62.20 (-92.30\%) & 0.11 & 5.20 \\

\midrule 
\multirow{7}{*}{5} & 
  1 & 4.09 & 2.74 & 218.74 & 527.52 & 1.09 & 1.09 \\
& 2 & 4.08 (0.24\%) & 2.64 (3.65\%) & 252.84 (15.59\%) & 761.86 (44.42\%) & 0.61 & 1.21 \\
& 3 & 4.11 (-0.49\%) & 2.57 (6.20\%) & 255.33 (16.73\%) & \textbf{848.14 (60.78\%)} & 0.44 & 1.32 \\
& 5 & 4.07 (0.49\%) & 2.56 (6.57\%) & \textbf{261.46 (19.53\%)} & 734.32 (39.20\%) & 0.31 & 1.55 \\
& 10 & 4.08 (0.24\%) & 2.51 (8.39\%) & 177.86 (-18.69\%) & 248.72 (-52.85\%) & 0.21 & 2.06 \\
& 20 & 4.07 (0.49\%) & 2.48 (9.49\%) & 77.82 (-64.42\%) & 93.50 (-82.28\%) & 0.15 & 3.01 \\
& 50 & 4.08 (0.24\%) & \textbf{2.44 (10.95\%)} & 19.56 (-91.06\%) & 29.24 (-94.46\%) & 0.12 & 5.53 \\

\midrule     
\multirow{7}{*}{10} &
  1 & 4.07 & 2.15 & 124.53 & 367.76 & 1.17 & 1.17 \\
& 2 & 4.08 (-0.25\%) & 2.07 (3.72\%) & \textbf{202.67 (62.75\%)} & \textbf{494.83 (34.55\%)} & 0.65 & 1.30 \\
& 3 & 4.09 (-0.49\%) & 2.02 (6.05\%) & 187.53 (50.59\%) & 487.43 (32.54\%) & 0.48 & 1.42 \\
& 5 & 4.09 (-0.49\%) & 2 (6.98\%) & 160.05 (28.52\%) & 283.30 (-22.97\%) & 0.34 & 1.68 \\
& 10 & 4.08 (-0.25\%) & 1.97 (8.37\%) & 102.65 (-17.57\%) & 118.40 (-67.81\%) & 0.23 & 2.22 \\
& 20 & 4.07 (-0.00\%) & 1.94 (9.77\%) & 53.80 (-56.80\%) & 55.36 (-84.95\%) & 0.16 & 3.23 \\
& 50 & 4.08 (-0.25\%) & \textbf{1.91 (11.16\%)} & 13.29 (-89.33\%) & 14.55 (-96.04\%) & 0.12 & 5.78 \\

\bottomrule
    \end{tabular}
\vspace{-0.3cm}
\end{table*}
}

\title{A Token-Wise Beam Search Algorithm for RNN-T}
\name{Gil Keren}
\address{Meta AI}
\copyrightnotice{979-8-3503-0689-7/23/\$31.00~\copyright2023 IEEE}

\begin{document}
%\ninept
%
\maketitle
\begin{abstract}
Standard Recurrent Neural Network Transducers (RNN-T) decoding algorithms for speech recognition are iterating over the time axis, such that one time step is decoded before moving on to the next time step. Those algorithms result in a large number of calls to the joint network, which were shown in previous work to be an important factor that reduces decoding speed. We present a decoding beam search algorithm that batches the joint network calls across a segment of time steps, which results in 20\%-96\% decoding speedups consistently across all models and settings experimented with. In addition, aggregating emission probabilities over a segment may be seen as a better approximation to finding the most likely model output, causing our algorithm to improve oracle word error rate by up to 11\% relative as the segment size increases, and to slightly improve general word error rate. 
\end{abstract}

\section{Introduction} \label{sec:intro}

% RNNTs are most common, but decoding is still much slower than CTC.
The Recurrent Neural Network Transducer model (RNN-T) \cite{graves2012sequence,graves2014towards} has become a common model of choice for on-device speech recognition over the past few years due to its attractive accuracy to model size tradeoff \cite{he2019streaming,li2020towards}. Indeed, compared to Connectionist Temporal Classification (CTC) models, RNN-T models normally yield a better word error rate (WER) for a fixed model size \cite{chiu2019comparison}. However, CTC models are still largely superior in decoding speed \cite{higuchi2021comparative}, which results from their conditional independence assumption.

% Previous work showed the number of joiner calls is the problem
During RNN-T decoding with standard algorithms, frames (time steps of the encoder network output) are being iterated over and decoded sequentially. For each frame, the joint network (the joiner) is applied to combine the encoder and prediction networks outputs, before deciding which tokens are added to the N-best hypotheses beam \cite{graves2012sequence,tripathi2019monotonic,boyer2021study}. The process repeats until no new tokens are emitted in this frame, and only then the next frame is decoded. Previous work has shown that even with a single layer joiner, its repeated application is responsible for a significant amount of decoding time and power consumption \cite{le2022factorized,saon2020alignment}, due to lack of parallelization and repeatedly loading the joiner weight matrix to memory. Note that for models using larger subword units, the joiner is being applied much more often than the prediction network, as the latter is only being applied when a new token is emitted. 

% Current RNNT does not batch joiner calls across frames. This fits a strict streaming scenario, but in practice streaming is done in non-trivial segments. So the goal is to batch joiner calls across frames in the segment. This fits both full sequences and streaming cases
While decoding in a strictly monotonic manner along the time axis may be simpler, and mandatory in strict streaming use cases, in practice this is often not required. For non-streaming cases, decoding algorithms may indeed benefit from the available future frames for better accuracy and computational efficiency. In practical streaming cases, streaming is normally done in a segment-by-segment manner, where each segment contains a few frames \cite{shi2021emformer,le2021deep}. Therefore in those scenarios as well, some lookahead is available for the decoding algorithm to benefit from. 

% Our algorithm explained
In this work, we present a novel beam search decoding algorithm for RNN-T that decodes audio utterances segment by segment. Standard RNN-T decoding algorithms attempt to find the highest scoring decoding paths in a greedy frame-by-frame manner, and the search space is trimmed to contain the highest scoring hypotheses after each frame or token emission. 
Instead of iterating over frames, our algorithm considers an entire segment of frames simultaneously, and aggregates probabilities of paths leading to identical sequences of token emissions within the entire segment. Our algorithm emits a token after considering all frames in the segment, and the search space is trimmed after each token emission or when the entire segment processing is done.
Note that in the extreme case, the segment can be the entire sequence. Since the algorithm does not iterate over frames within the segment anymore, we name this logic token-wise decoding. When using a segment size of one frame, our algorithm is identical to a standard breadth-first RNN-T decoding algorithm \cite{tripathi2019monotonic}. 

% The two benefits of this approach
Our decoding approach has two advantages. First, considering the entire segment simultaneously allows us to batch the joint network applications across multiple frames and reduce the total number of invocations of the joint network, which considerably improves the computational efficiency of the inference process. Second, token emission probabilities are potentially spread across a number of frames, which may cause a standard strictly monotonic decoding algorithm to discard a correct hypothesis during search. Our proposed algorithm aggregates emission probabilities across all frames of the decoded segment, which may improve decoding accuracy. This may be viewed as a theoretically motivated improvement to the heuristic search process, that provides a better approximation to finding the most probable token sequence. The algorithm does not require any change to RNN-T model training.

Note that while we reduce the total number of joiner calls, which in general improves decoding speed, each call incurs increased computation, since we are joining the entire segment instead of only one frame. This would not make sense for very long segments, as the total computation may increase dramatically. 
%Naturally, there is no need to process frames at the end of the segment, in order to find the most likely first token to emit in the segment. 
In the experiments below, we explore multiple segment sizes to find that segments containing about 3-5 frames (after the encoder stride) result in best decoding speed. 

% Overall summary, gains
Overall, our novel token-wise beam search algorithm for RNN-T generalizes a standard beam search algorithm, and employs a larger lookahead which is normally available in both streaming and non-streaming applications, to improve decoding speed and accuracy. In experiments using both CPUs and GPUs, we find that when using a segment size of 3-5 time frames, we manage to consistently increase the decoding speed by about 20\%-96\%. In addition, as the segment size increases, decoding accuracy improves, which is demonstrated by up to a 11\% improvement in oracle WER and a slight improvements in general WER.

\section{Related work}
% Other methods reducing joiner calls: monotonic RNNT, Duc's thing. Those can be combined with that. 
% another benefit from future looking: old german paper
We mention additional related work that was not otherwise mentioned in Section \ref{sec:intro}. Monotonic RNN-T \cite{tripathi2019monotonic} is another technique capable of reducing the number of joiner invocations. With Monotonic RNN-T, the loss function is altered such that every token emission results in advancing to the next time step, therefore each frame is joined at most once. However, without using complex initialization techniques, monotonic RNN-T normally results in some WER degradation compared to standard RNN-T \cite{moritz2023investigation}. Nevertheless, Our proposed algorithm can be easily adjusted to support monotonic RNN-T in order to speed up its decoding even further. 

Alignment-length synchronous decoding (ALSD) \cite{saon2020alignment} shows a reduction in joiner invocations and a decoding speedup, however this was compared to an algorithm forcing each hypotheses to a fixed number (at least 4) of expansions per frame. ALSD speedups come from the algorithm's ability to trim hypotheses early without expanding them, an ability the standard baseline algorithm considered in this work possesses as well. In addition, ALSD allows different hypotheses to be joined with different encoder frames, which cannot be done trivially using a single joiner call, therefore may result in a slowdown compared to other optimized implementations.  

Another technique reducing the number of joiner invocations is presented in \cite{le2022factorized}, by using HAT factorization \cite{variani2020hybrid} and a threshold on the blank emission probability, above which the joiner will not be invoked. While this method requires replacing standard RNN-T with the HAT variant and finding the appropriate threshold to use, our proposed algorithm can be altered to benefit from this method as well, such that speed gains from both methods are cumulative.

\section{Method} \label{sec:method}
% Define RNNT again
\subsection{Recurrent Neural Network Transducer}
Consider an audio segment (possibly features) denoted $x$, and a sequence of reference tokens (during training) or previously emitted tokens (during inference) $y=(y_1, ..., y_U)$.
The RNN-T model has three main components. The encoder network, processing an audio segment with possible downsampling to length $T$: $h(t) = \mathtt{encoder}(x)$ , the prediction network processing the reference or emitted tokens: $g(u) = \mathtt{predictor}(y)$, and the joint network (the joiner) combining the above into token scores: $j(t, u, k) = \mathtt{joiner}(h, g)$. Here $k$ denotes the token from a vocabulary $V \cup \{\phi\}$ where $\phi$ is the blank symbol.

During decoding with an RNN-T model, the search process aims at finding the most probable output sequence: $y^* = argmax _{y} \, p(y|x)$.
Solving for the latter is intractable in practice, therefore beam search algorithms are normally used. Common versions are variants of the original algorithm \cite{graves2012sequence}, such as \cite{tripathi2019monotonic,boyer2021study,saon2020alignment}. In this work, we refer to the breadth-first beam search algorithm from \cite{tripathi2019monotonic} as the standard decoding algorithm. This standard algorithm was used in many works including \cite{prabhavalkar2021less,mahadeokar2021alignment,le2021deep,jain2020contextual,xiao2021scaling}.

The standard algorithm decodes along the time axis of the audio. At each iteration, given $N$ hypotheses in the beam, the joiner computes token emission probabilities for the current frame, including blank emission. Existing hypotheses are then expanded by either a blank or a non-blank symbol:
\begin{equation}
\label{eq:standard}
    p(y_1, ..., y_U, y_{U+1}) = p(y_1, ..., y_U) \cdot J(t, U, y_{U+1})
\end{equation}
where $J$ is the joiner output $j$ after softmax normalization, and $y_{U+1} \in V \cup \{\phi\}$ here.
%can be either blank or non-blank here. 
The most likely $N$ new hypotheses replace the previous ones in the beam. The probability of different alignments corresponding to the same token sequence are summed. Since RNN-T allows multiple token emissions in a single time-step, the next iteration moves to the next frame only after all hypotheses in the beam end with a blank symbol emission. As a result, the joiner is invoked at least once per frame, and at least once more if some non-blank tokens are emitted in any of the $N$ hypotheses in the beam during this frame.

\subsection{Token-Wise Beam Search}
As motivated above, our goal is to speed up decoding by reducing the total number of joiner invocations. Decoding is often done in an offline manner for an entire sequence, or in a non-strict streaming setting in a segment-by-segment manner, with each segment containing multiple encoder output frames. Therefore, we design an algorithm that decodes a number of frames simultaneously, allowing batching joiner invocations across frames and potentially improving the search accuracy. 

\alghere

\tabresone

\tabrestwo

The standard algorithm iterates over frames, and for each frame finds the highest scoring hypothesis expansions. The search space is normally trimmed to contain the N-best hypotheses after each token emission and before the next frame is processed.  
Instead, we consider the entire segment of frames simultaneously, and aggregate probabilities of paths leading to identical sequences of token emissions within
the entire segment. 
Note that in this case, the emission of the token is not associated anymore with any specific frame in the segment, but it results in a distribution over the segment frames that describes where the token was emitted. This distribution is used in the next token emission iteration, to continue aggregating the path probabilities across a segment correctly and to make sure the next token cannot be emitted before the previous one. 

%Our algorithm operates in a token-wise manner. Given a segment of $S$ frames to decode, we sum path probabilities across the segment to find the most likely tokens to expand the current $N$ hypotheses. Those expansions may happen in any frame, and not necessarily at the first frame. We continue by searching again for the most likely next token emissions across the sequence, making sure the second token is emitted after the first one. The search continues until no non-blank tokens emissions for the sequence are in the beam. 

We describe mathematically how to aggregate emission probabilities across a sequence of frames of length $S$. While the mathematics is a bit more complex than for the standard algorithms, it only amounts to summing the probabilities of the correct paths within a segment that lead to a given sequence of tokens. 
Assume a prefix of non-blank tokens $y = (y_1, ..., y_U)$ with probability $p(y)$. For each frame $t_1$ denote the associated probability $p(y_1, ..., y_U(t_1))$ of the sequence $y$ where the last token $y_U$ was emitted in frame $t_1$, such that:
\begin{equation}
    p(y) = \sum _{t_1=1}^S p(y_1, ..., y_U(t_1)).
\end{equation}
Define the probability of emitting blank from frame $t_1$ to frame $t_2$ as:
\begin{equation}
\label{eq:eq3}
Bl(U, t_1, t_2) = 
\begin{cases}
    \prod _{t'=t_1}^{t_2 - 1} J(t', U, \phi) & t_1 < t_2 \\
    1 & t_1 = t_2 \\
    0 & t_1 > t_2.
\end{cases}
\end{equation}
The probability to emit a new token $y_{U+1}$ at frame $t_2$, given that the previous token was emitted in an earlier frame $t_1$, is the probability of emitting blanks from $t_1$ to $t_2$ and emit $y_{U+1}$ at $t_2$:
\begin{multline}
\label{eq:eq4}
p(y_1, ..., y_U(t_1), y_{U+1}(t_2)) = \\
p(y_1, ..., y_U(t_1)) \cdot Bl(U, t_1, t_2) \cdot J(t_2, U, y_{U+1}).
\end{multline}
The probability to emit a new token $y_{U+1}$ at time $t_2$ is then factored across the different $t_1$ locations:
\begin{multline}
\label{eq:eq5}
 d = p(y_1, ..., y_U, y_{U+1}(t_2)) = \\ 
\sum _{t_1=1}^S p(y_1, ..., y_U(t_1), y_{U+1}(t_2)).
\end{multline}
Finally, the probability of expanding $y$ with a the new non-blank token $y_{U+1}$ across the segment is then factored across the different $t_2$ locations:
\begin{multline}
\label{eq:eq6}
    \delta^{nb} = p(y_1, ..., y_{U+1}) = \sum _{t_2=1}^S  p(y_1, ..., y_U, y_{U+1}(t_2)).
\end{multline}
For blank expansions, the token sequence does not change, and its probability is updated using blank emission probabilities for the rest of the segment:
\begin{multline}
\label{eq:eq7}
\delta^\phi = p(y) = \sum _{t_1 = 1}^S p(y_1, ..., y_U(t_1)) \cdot Bl(U, t_1, S+1).
\end{multline}

\tabresthree

Equations \ref{eq:eq3}-\ref{eq:eq7} are used to compute the aggregated token emission probabilities across the segment. When starting to decode a segment, we assume the distribution of $p(y_1, ..., y_U(t_1))$ is concentrated in the first frame of the segment ($t_1 = 1)$. In consecutive steps, Eq. \ref{eq:eq5} from the previous step is used as the distribution over last token emission frames. The rest of the algorithm is identical to the standard algorithm, including summing scores of hypotheses that correspond to the same token sequence. 
Note that when $S=1$, the sequence expansion probabilities amount to precisely the way those are computed in the standard algorithm time-wise algorithm in Eq. \ref{eq:standard}, therefore the proposed algorithm is a natural generalization of the standard algorithm. 

The full algorithm is given at Algorithm \ref{alg:alg1}, with all parts other than Equations \ref{eq:eq3}-\ref{eq:eq7} being fairly standard. On a high level, at every iteration, a segment is processed, and all existing hypotheses start in group A. Token scores are computed using the above equations, are top scoring tokens are added to expand the hypotheses. When a blank token is added, the hypothesis is moved to group B and is considered done for the current segment. When no more hypotheses remain in group A, we move to the next segment.

Note that for simplicity, the algorithm assumes the encoder output was computed beforehand. In streaming applications, one may want to compute the encoder output in a segment-by-segment manner as well, and integrate that into Algorithm \ref{alg:alg1}. The segment size used for encoder computation and decoding are not necessarily constraint to be equal. In initial experiments, we included a version of the algorithm that limits the maximum number of tokens to emit in a segment, similar to the N-step constrained beam search from \cite{boyer2021study}. However, we found that for all segment sizes, tuning this hyperparameter does not lead to any compute savings without a considerable degradation in WER, therefore we omit this functionality from the algorithm. 

We briefly explain some of the simple helper functions used in Algorithm \ref{alg:alg1}. The function $\mathtt{InitPredictor}$ returns the initial state of the predictor, which is initialized the same way as a weight matrix. $\mathtt{Length}$/$\mathtt{Append}$ are the standard list operations. $\mathtt{ChooseNBest}$ returns the $N$ hypotheses with highest scores out of a list of hypotheses. $\mathtt{AddAndMerge}$ adds a hypothesis to a list of hypotheses, and in case a hypothesis with the same tokens already exists in the list, the two are merged by summing their scores. $\mathtt{ChooseNthScore}$ returns the $N$-th highest score of a hypotheses list. $\mathtt{ChooseNBestExpansions}$ takes the non-blank expansion scores for all hypotheses and returns the top $N$ expansions by score as pairs comprising of a hypothesis index and an expansion token. 
$\mathtt{UpdatePredictorOutput}$ runs the RNN-T predictor with the updated hypotheses tokens, and updates their predictor output member.

\vspace{-0.3cm}
\section{Experiments}
\subsection{RNN-T Models and Data}
Three different standard RNN-T models are evaluated. The first is trained on Librispeech data \cite{panayotov2015librispeech}, where the encoder network contains 20 Emformer layers \cite{shi2021emformer} and performs a total stride of 4, the prediction network contains 3 LSTM layers, and the joiner is a ReLU layer followed by a single fully connected layer projecting the representation to the vocabulary dimension. For this model, 500 subword units are used \cite{kudo2018sentencepiece}. This model has in total of 78M parameters. 
% Gil: versions
%The second and third models are trained on large in-house dataset containing 1.5M hours of videos and voice assistant data. This dataset is annotated partly by humans and partly by existing ASR systems. Further details about the in-house datasets are omitted to support the double-blind review process. 
The second and third models are trained on large in-house dataset containing 1.5M hours speech data. Our in-house training set combines two sources. The first consists of English video data publicly shared by Facebook users; all videos are completely de-identified. The second contains de-identified English data with no user-identifiable information in the voice assistant domain. All utterances are morphed when the speakers are de-identified. Note that the data is not morphed during training. 
The second model contains 28 Emformer layers in the encoder network with a total stride of 4, and 2 LSTM layers in the prediction network. The third model contains 13 Emformer layers with a total stride of 6, and a single LSTM layer in the prediction network. The joint network is identical to the ones used in the Librispeech model. The total number of parameters in the second and third models are 104M and 27M respectively. We evaluate the second model on a voice assistant test set, and the third model on a voice dictation test set.

\vspace{-0.3cm}
\subsection{Results}
We evaluate the token-wise beam search algorithm using different segment sizes to measure accuracy and decoding speed. When using a segment size of one, our algorithm is identical to the standard algorithm, as also verified in initial experiments. Therefore all results are reported using the same implementation, to avoid any noise during performance measurements. Results on three datasets appear in Tables \ref{tab:tab1}, \ref{tab:tab2} and  \ref{tab:tab3}. 
% Gil: versions
% Due to space limitations, additional results are given in the supplementary material, which follow a similar trend. 
The results in the appendix follow a similar trend. 
The CPU / GPU results in the above mentioned tables were obtained using a single core CPU / single GPU. 

The first observation from the results is that using the token-wise beam search with optimized segment size of 2-5 frames consistently improves throughput over the standard algorithm (segment size of 1). Throughput is measured as number frames (of the encoder output) decoded per second. The best segment size results in 20\%-96\% throughput increase over the standard algorithm. The last two columns provide further insight regarding the speed gains. Calls / Frame is the average number of times the joint network is invoked per frame, which decreases consistently as the segment size increases. On the other hand, each joiner call involves a larger number of frames as the segment size increases. The Joins / Frame column measures the total joiner computation as the average number each frame was in the joiner input. Overall, total compute increases with the segment size, but since the number of joiner calls decreases at the same time and each call incurs an extra cost, segment sizes of 3-5 result in a good tradeoff with considerable overall speedups. 

The second observation is that token-wise beam search benefits from aggregating emission probabilities across a segment, as seen in improved Oracle WER (OWER) as the segment size increases. Oracle WER measures the maximum WER in the $N$ best hypotheses returned by the search algorithm. Improvements in OWER are up to 11\% relative for a segment size of 50 frames. As the entire set of $N$ best hypotheses are often processed by downstream applications such as natural language understanding models \cite{liu2021asr,li2020improving,weng2020joint,deoras2012joint}, this improvement may be useful in those situations. 

\vspace{-0.3cm}
\section{Conclusion}
\vspace{-0.3cm}
We proposed a token-wise beam search algorithm for RNN-T models that does not require any changes to the trained model, and can be applied in any offline or standard non-strict streaming decoding setting. Our algorithm aggregates emission probabilities over segments of frames, thus reducing the number of joiner invocations, which result in consistent 20\%-96\% speedups, and improves search accuracy as seen in up to 11\% oracle WER improvement. In future work, we plan to adjust this algorithm to support monotonic RNN-T \cite{tripathi2019monotonic} and blank thresholding \cite{le2022factorized} to further improve decoding speed.

\bibliographystyle{IEEEbib}
\bibliography{strings,refs}

\clearpage

\iftrue{

\appendix
\section{Additional results}
See pages below for additional result tables.

\begin{table*}[htb]
  \caption{Librispeech model (test-clean): decoding results with the standard algorithm (segment size 1) and our proposed token-wise beam search. In parentheses are the relative improvement \% over the standard algorithm. Best throughput and OWER values are in boldface. \vspace{0.1cm}}
  \label{tab:tab4}
  \centering
  \begin{tabular}{ c c c c c c c c}
    \toprule
        N-best & Segment & WER & OWER & Throughput  & Throughput & Calls/ & Joins/ \\
               & Size    &     &      &  CPU (f/s)  & GPU (f/s)  & Frame  & Frame \\

\midrule 
\multirow{7}{*}{1} & 
  1 & 3.42 & 3.42 & 284.74 & 877.55 & 1.13 & 1.13 \\
& 2 & 3.39 (0.88\%) & 3.39 (0.88\%) & 407.89 (43.25\%) & 1,251.11 (42.57\%) & 0.64 & 1.27 \\
& 3 & 3.4 (0.58\%) & 3.4 (0.58\%) & \textbf{552.30 (93.97\%)} & \textbf{1,411.64 (60.86\%)} & 0.47 & 1.40 \\
& 5 & 3.4 (0.58\%) & 3.4 (0.58\%) & 413.66 (45.28\%) & 1,381.28 (57.40\%) & 0.34 & 1.67 \\
& 10 & 3.38 (1.17\%) & 3.38 (1.17\%) & 342.11 (20.15\%) & 1,081.46 (23.24\%) & 0.24 & 2.34 \\
& 20 & 3.36 (1.75\%) & \textbf{3.36 (1.75\%)} & 186.54 (-34.49\%) & 257.02 (-70.71\%) & 0.19 & 3.64 \\
& 50 & 3.37 (1.46\%) & 3.37 (1.46\%) & 57.70 (-79.74\%) & 67.73 (-92.28\%) & 0.16 & 7.40 \\

\midrule     
\multirow{7}{*}{2} & 
  1 & 3.32 & 3.01 & 182.71 & 453.58 & 1.43 & 1.43 \\
& 2 & 3.33 (-0.30\%) & 2.86 (4.98\%) & 199.65 (9.27\%) & 734.15 (61.86\%) & 0.76 & 1.53 \\
& 3 & 3.33 (-0.30\%) & 2.79 (7.31\%) & \textbf{288.59 (57.95\%)} & 851.07 (87.63\%) & 0.55 & 1.65 \\
& 5 & 3.33 (-0.30\%) & 2.68 (10.96\%) & 229.88 (25.82\%) & \textbf{886.42 (95.43\%)} & 0.39 & 1.93 \\
& 10 & 3.3 (0.60\%) & 2.62 (12.96\%) & 150.92 (-17.40\%) & 412.78 (-9.00\%) & 0.27 & 2.63 \\
& 20 & 3.31 (0.30\%) & 2.54 (15.61\%) & 107.76 (-41.02\%) & 130.68 (-71.19\%) & 0.20 & 3.94 \\
& 50 & 3.3 (0.60\%) & \textbf{2.48 (17.61\%)} & 28.51 (-84.40\%) & 30.79 (-93.21\%) & 0.16 & 7.66 \\

\midrule 
\multirow{7}{*}{5} & 
  1 & 3.31 & 2.04 & 76.81 & 287.40 & 1.60 & 1.60 \\
& 2 & 3.32 (-0.30\%) & 1.96 (3.92\%) & 97.17 (26.51\%) & 429.32 (49.38\%) & 0.84 & 1.68 \\
& 3 & 3.31 (-0.00\%) & 1.93 (5.39\%) & 121.88 (58.68\%) & \textbf{497.81 (73.21\%)} & 0.60 & 1.80 \\
& 5 & 3.3 (0.30\%) & 1.9 (6.86\%) & \textbf{130.72 (70.19\%)} & 390.02 (35.71\%) & 0.42 & 2.10 \\
& 10 & 3.29 (0.60\%) & 1.85 (9.31\%) & 71.24 (-7.25\%) & 139.43 (-51.49\%) & 0.29 & 2.84 \\
& 20 & 3.3 (0.30\%) & \textbf{1.82 (10.78\%)} & 39.28 (-48.86\%) & 59.35 (-79.35\%) & 0.22 & 4.20 \\
& 50 & 3.3 (0.30\%) & 1.84 (9.80\%) & 10.06 (-86.90\%) & 14.38 (-95.00\%) & 0.17 & 8.05 \\

\midrule 
\multirow{7}{*}{10} & 
  1 & 3.29 & 1.67 & 48.88 & 195.44 & 1.72 & 1.72 \\
& 2 & 3.31 (-0.61\%) & 1.61 (3.59\%) & 62.04 (26.92\%) & 281.52 (44.04\%) & 0.91 & 1.81 \\
& 3 & 3.29 (-0.00\%) & 1.59 (4.79\%) & \textbf{81.55 (66.84\%)} & \textbf{283.35 (44.98\%)} & 0.65 & 1.93 \\
& 5 & 3.3 (-0.30\%) & 1.58 (5.39\%) & 58.48 (19.64\%) & 168.04 (-14.02\%) & 0.45 & 2.23 \\
& 10 & 3.3 (-0.30\%) & 1.55 (7.19\%) & 54.44 (11.37\%) & 72.46 (-62.92\%) & 0.30 & 2.98 \\
& 20 & 3.3 (-0.30\%) & 1.54 (7.78\%) & 28.29 (-42.12\%) & 29.20 (-85.06\%) & 0.23 & 4.40 \\
& 50 & 3.29 (-0.00\%) & \textbf{1.52 (8.98\%)} & 5.17 (-89.42\%) & 7.12 (-96.36\%) & 0.18 & 8.34 \\

\bottomrule
    \end{tabular}
\end{table*}
}
\fi

\end{document}